# A Reduced Reference Image Quality Measure Using Bessel K Forms Model for Tetrolet Coefficients

[1]Abdelkaher Ait Abdelouahad, [2]Mohammed El Hassouni, [3]Hocine Cherifi, [4]Driss Aboutajdine

[*1, 4]*LRIT URAC 29-University of Mohammed V-Agdal-Morocco,*
A.abdelkaher@gmail.com
Aboutaj@fsr.ac.ma

[2,] *DESTEC FLSHR- University of Mohammed V-Agdal-Morocco,*
Mohamed.elhassouni@gmail.com

[3,]*Le2i-UMR CNRS 5158-University of Burgundy, Dijon-France,*
Hocine.cherifi@u-bourgogne.fr

**Abstract**

*In this paper, we introduce a Reduced Reference Image Quality Assessment (RRIQA) measure based on the natural image statistic approach. A new adaptive transform called "Tetrolet" is applied to both reference and distorted images. To model the marginal distribution of tetrolet coefficients Bessel K Forms (BKF) density is proposed. Estimating the parameters of this distribution allows to summarize the reference image with a small amount of side information. Five distortion measures based on the BKF parameters of the original and processed image are used to predict quality scores. A comparison between these measures is presented showing a good consistency with human judgment.*

## 1. Introduction

Since the advent of digital images, image processing techniques are increasingly in demand. However, such techniques like compression, quantization watermarking [1] and denoising [2] affect the image perceived quality. Thus, an objective image quality assessment measure is required to compare and to monitor the performances of image processing algorithms [3]. Depending on the availability of the reference image we can distinguish three types of objective image quality assessment methods. When the reference image is available the measures belong to the Full Reference (FR) class. The Peak Signal-to-Noise Ratio (PSNR) and the Mean Structural Similarity Index (MSSIM)[4], are both widely used FR methods. No reference (NR) methods, aim to quantify the quality of a distorted image without any cue from its original version. They are generally conceived for specific distortion type and cannot be generalized for other distortions [5]. Reduced Reference (RR) methods are typically used when one can send side information with the processed image related to the reference. Recently, a number of authors have successfully introduced RR methods based on: image distortion modeling [6][7], human visual system (HVS) modeling [8][9], and finally natural image statistics modeling [10][11]. As our work falls in the last approach we recall its underlying assumption. Natural images statistics are the basic stimuli that our visual system is adapted to. After processing, the statistics of the images change and make it unnatural. Understanding the way by which statistics change and measuring these changes allows us to predict the visual degradation.

In [10], Wang et al used the steerable pyramids to represent the distorted and the reference images in the spatial-frequency domain. First, the subbands coefficients are fitted with the Generalized Gaussian Density (GGD). Second, the Kullback Leibler Divergence (KLD) is used to quantify the visual degradation. This approach has introduced a convenient way to assess the quality. It provides quality scores highly correlated with human judgment for a wide range of distortion types. However, with only four orientations the steerable pyramids fails to span 360° of orientation space accurately.

Wufeng [12], proposed the use of eight orientations and six scales of the same transform. For each scale, the Strongest Component Map (SCM) is constructed from the coefficients with maximum amplitude among all orientations, and then the coefficients of the SCM are fitted to a Weibull distribution. The absolute deviation and the relative deviation are computed between the scale parameter of a distorted image and the scale parameter of the reference image. Finally, the summation of the geometric mean of these deviations is derived to quantify the perceptual distortion. Important improvements were remarked compared with the previous approach. These results motivate our proposition to use a more flexible image representation and to derive an adequate image statistic model. The paper is organized as follows: Section 2 relates our motivations and presents the general scheme of the proposed RRIQA. The tetrolet transform is presented in section 3. Section 4, describes tetrolet coefficients statistics while section 5 concerns the distortion measure. Section 6 relates our experiments and results. Finally, a conclusion ends the paper.

## 2. Method

Steerable Pyramids provide a convenient framework for localized representation of signals simultaneously in space and frequency. However, using this representation, the geometric structures of the image are disregarded. This can be explained by the nature of the basis functions of the steerable pyramids which are directional derivative operators. To solve this drawback we propose a new adaptive transform derived from the Haar wavelet transform and based on "teterominoes". Tetrominoes are shape obtained by connecting four equal sized squares. "Tiling by tetrominoes" is the key point of this transform, where an optimal tetrominoes-based covering is wanted. The covering is optimal in the sense that the local geometric structures are taken into account. Furthermore, with only three orientations the combination of tetrominoes in each covering provides more directions adapted to the local structural information in the image.

The tetrolet transform is a sparse representation which reduces the number of wavelet coefficients compared with the classical tensor product wavelet transform. Thus, the tetrolet coefficients distribution is highly non Gaussian with a heavy tails and a sharp peak. Such a behavior can involves the use of kurtosis as a measure of peakedness. In [13] and [14], NR measures have been implemented upon the kurtosis, as an index of quality. High correlation with HVS was obtained. However, in the context of RR, a measure based on kurtosis cannot always reflect the visual degradation. This is due to the large magnitude of the kurtosis in some subbands. One can use a probability density function (PDF) to model marginal distribution of tetrolet coefficients. Wang et al [10], have used the Generalized Gaussian Density (GGD) for modeling steerable pyramids coefficients. Here, we propose the BKF density for two reasons : 1) it is a leptokurtic distribution, and as it was shown in [15], the BKF can well fit the observed marginal distribution comparing to Symmetric Alpha Stable (SαS) [16] and GGD models, 2) its parameters are estimated essentially based on the kurtosis [17], thus a measure based on the BKF parameters takes into account the peakedness of tetrolet coefficients.

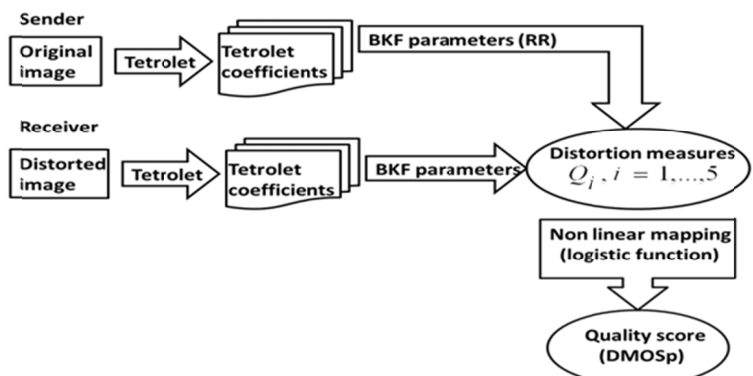

**Figure 1.** The deployment scheme of the proposed RRIQA measure

Once the BKF parameters are estimated, we proceed to calculate the quality measures. First, we introduce measures which use either the shape parameter or the scale parameter of the BKF density. This can help to reduce the number of features extracted from the reference image. These measures are based on the absolute value of the difference between parameters or are based on the absolute and the relative deviations between those ones. Second, we present a measure which incorporates the two parameters together. This later uses the pseudo metric to quantify the difference between two BKF probability density functions (PDFs). Fig.1 illustrates the RRIQA proposed scheme. This scheme provides a measure between a reference image and its distorted version.

## 3. Tetrolet transform

The Haar wavelet transform consist in averaging of sum and differences of pixels arranged in $2\times 2$ squares, this necessitates a dyadic partition of the image index set $E = \left\{ I_{0,0},...,I_{\frac{N}{2},\frac{N}{2}} \right\}$ where for $i,j = 0,...,\frac{N}{2}$, we have $I_{i,j} = \{(2i,2j),(2i+1,2j),(2i,2j+1),(2i+1,2j+1)\}$. Although, this square based partition takes into account structural image. However image local geometry is disregarded. Figure 2 shows an example where the Haar transform is not able to capture the local geometry of the image. The idea of tetrolet transform is to allow more general partitions which capture the image local geometry by bringing the "tiling by tetrominoes" problem into play.

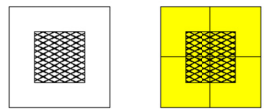

**Figure 2.** The Haar transform applied to an image block.

Tetrominoes are derived from the well know game "tetris". They were introduced by Golomb [18]. We can obtain a tetromino by connecting four equal sized square. Disregarding rotation and isometric we have five free tetrominoes as shown in Figure 3.

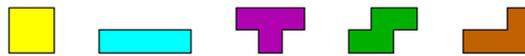

**Figure 3.** the five free tetrominoes

The Haar transform is a special case, since it considers only the first tetromino (square). To use other tetrominoes we should have at least a $4\times 4$ blocks (Figure 4) which will give 117 possibility, whereas $8\times 8$ blocks gives $117^4 > 10^8$ possibilities. From computational complexity standpoint, it's clear that the first choice is the reasonable one.

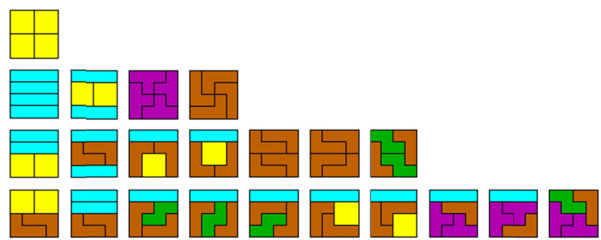

**Figure 4.** The 22 fundamental forms tiling a 4x4 board.

Let's take the example in Figure 2, several tetrominoes-based covering are possible, the optimal one is the one whose tetrominoes do not intersect an important geometric structure like an edge. Such optimal covering is illustrated in Figure 5. Therefore, tetrominoes ensure more directions when rotations and reflections are considered. To illustrate this let's take from Figure 4 (Line 4) the third covering (from left to right), eight other covering are possible with different directions are shown in Figure 6.

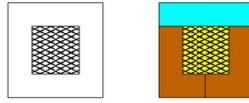

**Figure 5.** An optimal tiling for the proposed image block (left).

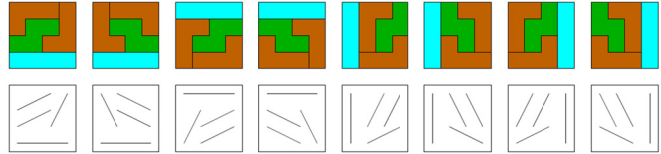

**Figure 6.** Different directions covered by the same tetrominoes.

The adaptivity of the tetrolet transform to the local geometric structures comes from the optimization process used in the filter bank algorithm [19]. Given a $4 \times 4$ block, the 117 admissible covering are considered. For each covering the Haar transform is applied to its tetrominoes, this generates four low-pass coefficients and 12 tetrolet coefficients, then the $l^1$ norm of the 12 tetrolet coefficients is used to determine the optimal covering. This later is chosen as the one who have the minimal $l^1$ norm. In other words, the smaller is the magnitude of the 12 tetrolet coefficients, the minimal is the $l^1$ norm. Thus we obtain the optimal covering and a sparse image representation.

## 4. Tetrolet coefficients statistics

The strong adaptive relevance of tetrolet transform and its sparseness provide good reasons to use it in IQA purpose. Moreover, we can demonstrate the sensitivity of tetrolets coefficients to various distortion types. For this goal, we apply a tetrolet transform to different distorted images and we observe how these distortions change the statistics of tetrolet coefficients. Figure.7 illustrates the results of this study. Figure .7 (a) represents a Gaussian blurred image and the histogram of tetrolet coefficients of a subband of this image (solid curve). The histogram of the same subband of the original image is also reported for comparative purpose (dashed curve). Figure .7 (b) and Figure .7 (c) give the same representations for the white noise contaminated image and the transmission errors distorted image, respectively.

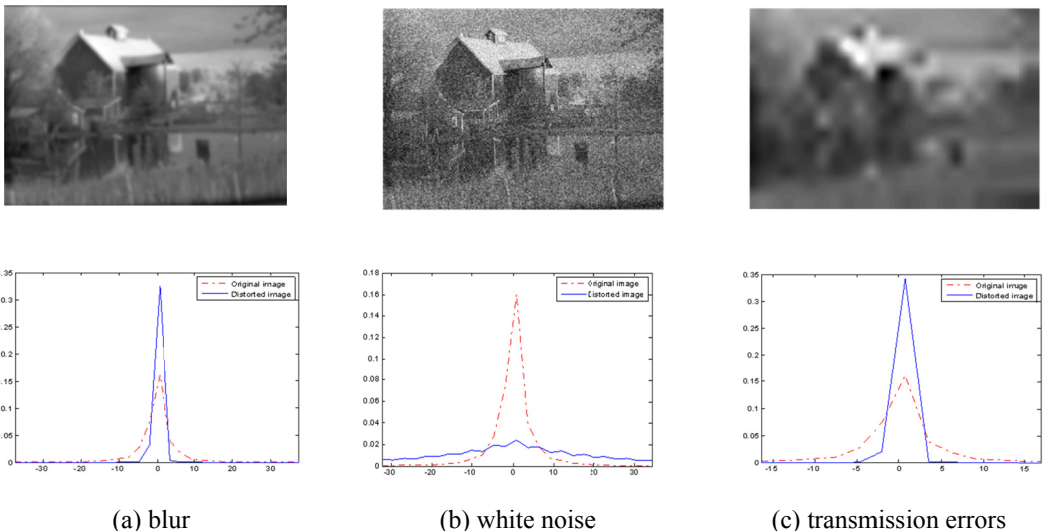

    (a) blur     (b) white noise     (c) transmission errors
**Figure 7.** Distorded Images and corresponding Histograms of Tetrolet coefficients under various distortion types. Dashed curves : the original image. Solid curves : distorted image.

The PDF of the BKF model is given by [15] :

$$f(x;\alpha,\beta) = \frac{1}{\sqrt{\pi}\Gamma(\alpha)}\left(\frac{\beta}{2}\right)^{-\frac{\alpha}{2}-\frac{1}{4}}\left|\frac{x}{2}\right|^{\alpha-\frac{1}{2}} K_{\alpha-\frac{1}{2}}\left(\sqrt{\frac{2}{\beta}}|x|\right), \quad (1)$$

Where $\Gamma(.)$ is the Gamma function, $\alpha$ and $\beta$ are the shape and scale parameters, respectively. $K_\nu$ is the modified Bessel function given by:

$$K_\nu(zx) = \frac{\Gamma\left(\nu+\frac{1}{2}\right)(2x)^\nu}{\sqrt{\pi}z^\nu}\int_0^\infty \frac{\cos(zt)}{(t^2+x^2)^{\nu+\frac{1}{2}}}dt, \quad (2)$$

The BKF parameters can be estimated using the Maximum Likelihood method [17] :

$$\hat{\alpha} = \frac{3}{kurtosis(x)-3} \quad \text{and} \quad \hat{\beta} = \frac{variance(x)}{\hat{\alpha}}$$

## 5. Distortion measures

The first two measures are based on the absolute value of the difference between the shape or scale parameter of a subband and its corresponding from the distorted image. For the rest of this section we consider $\alpha_r$ ($\beta_r$) and $\alpha_d$ ($\beta_d$) as the shape (scale) parameter of a subband from the reference image and its corresponding from the distorted image, respectively. The measures are:

$$Q_1 = \sum_{all\ bands} |\alpha_r - \alpha_d|, \quad (3)$$

$$Q_2 = \sum_{all\ bands} |\beta_r - \beta_d| \quad (4)$$

The second type of measure is based on absolute deviation $A_d^\alpha$ and relative deviation $R_d^r$ [12]. For the shape parameter, we have :

$$A_d^\alpha = |\alpha_r - \alpha_d| \quad \text{and} \quad R_d^r = \frac{|\alpha_r - \alpha_d|}{\alpha_r}$$

The summation of the geometric mean of $A_d^\alpha$ and $R_d^r$ over all Subbands is computed to predict the visual degradation as follows :

$$Q_3 = \sum_{all\ bands} \sqrt{A_d^\alpha \times R_d^\alpha}, \quad (5)$$

The same measure can be derived for the scale parameter β as follows :

$$Q_4 = \sum_{all\ bands} \sqrt{A_d^\beta \times R_d^\beta}, \quad (6)$$

To quantify the difference between two PDFs the well known Kullback Leibler Divergence (KLD) can be used. Nevertheless, according to our knowledge there is no closed form for the KLD for BKF PDFs. Here, we propose the use of $L^2$ metric [15]. Let $f(x;\alpha_1,\beta_1)$ and $f(x;\alpha_2,\beta_2)$ be two BKF PDFs, the $L^2$ metric can be expressed as :

$$d(\alpha_1,\beta_1,\alpha_2,\beta_2) = \sqrt{\int_x \left(f(x;\alpha_1,\beta_1) - f(x;\alpha_2,\beta_2)\right)^2 dx} \quad (7)$$

More explicitly, this distance can be simplified as :

$$d(\alpha_1,\beta_1,\alpha_2,\beta_2) = \left(\frac{1}{2\sqrt{2\pi}}\Gamma(0.5)\left(\frac{\xi(2\alpha_1)}{\sqrt{\beta_1}} + \frac{\xi(2\alpha_2)}{\sqrt{\beta_2}} - \frac{2\xi(\alpha_1+\alpha_2)}{\sqrt{\beta_1}}\left(\frac{\beta_1}{\beta_2}\right)^{\alpha_2}\mathcal{F}\right)\right)^{\frac{1}{2}}, \quad (8)$$

Where $\xi(\alpha) = \frac{\Gamma(\alpha+0.5)}{\Gamma(\alpha)}$ and $\mathcal{F} = F\left((\alpha_1+\alpha_2-0.5),\alpha_2;\alpha_1+\alpha_2,1-\frac{\beta_1}{\beta_2}\right)$ is the hypergeometric function. Finally, the measure between a reference image and its distorted version is given by:

$$Q_5 = \sqrt{\int_x \left(\sum_{all\ bands} d(\alpha_1,\beta_1,\alpha_2,\beta_2)^2\right)dx}, \quad (9)$$

## 6. Experimental results

Our experimental test was carried out using the LIVE database [20]. It is constructed from 29 high resolution images and contains seven sets of distorted and scored images, obtained by the use of five types of distortion at different levels. Set1 and 2 (labeled JP2 (1) and JP2(2)) are JPEG2000 compressed images, set 3 and 4 (labeled JPEG1 and JPEG2) are JPEG compressed images, set 5, 6 and 7 are respectively : Gaussian blur (Labeled Blur) , white noise ( labeled Noise) and transmission errors distorted images (Labeled Error). The 29 reference images have very different textural characteristics, various percentages of homogeneous regions, edges and details. The tests consist in choosing a reference image and one if its distorted versions; those later are considered as entries of the scheme given in Fig.1. First, a tetrolet transform with three levels is applied to both images, followed by a feature extraction step. Since tetrolet transform is based on the Haar transform, we have three orientations. This leads to 9 subbands (3 levels × 3 orientations). From each subband we extract the BKF parameters, so we obtain a vector of 18 features (9 subbands × 2 parameters). Before the transmission , these features are quantized. More explicitly, both, the shape and the scale parameters are quantized into 8 bits. So 144 bits are used to represent the RR features. Second, each of the measures introduced in the previous section is computed between the reference and distorted images. Finally, The objective quality scores (DMOSp) are computed from the values generated by the distortion measures, using a non linear mapping function proposed by the Video Quality Expert Group (VQEG) Phase I FR-TV [21]. Here, we use a four parameter logistic function :

$$logistic(\gamma,Q) = \frac{(\gamma_1-\gamma_2)}{1+e^{-\left(\frac{Q-\gamma_3}{\gamma_4}\right)}} + \gamma_2 \quad \text{where } \gamma = (\gamma_1,\gamma_2,\gamma_3,\gamma_4), \text{ then } DMOS_p = logistic(\gamma,Q)$$

The quality prediction performance of the considered objective metrics is quantified in terms of accuracy and monotonicity as recommended by the Video Quality Experts Group [21]. For comparative purpose we also report the performance of the WNSIM measure proposed in [10] which is actually the standard RR measure. We believe that such comparison is a fair one as the amount of side information necessary is comparable to the one used in the proposed measures. Indeed 72 bits are necessary to represent the RR features from $Q_1$ to $Q_4$, 144 bits for $Q_5$ while for the WNISM measure 162 bits are required. Prediction accuracy is quantified using the Pearson linear correlation coefficient while the prediction monotonicity is measured by the nonparametric Spearman rank order coefficient. This measure is used to quantify if changes (increase or decrease) in one variable is followed by changes (increase or decrease) in another variable, irrespective of the magnitude of the changes. Note that after preliminary fitting objective metric values and MOS, it is, in general, possible to apply the conventional Pearson correlation instead of rank correlation. Meanwhile, it is important to underline that the quality of fitting may reduce the accuracy of assessing a metric correspondence to HVS. Performances of the introduced measures are provided for all the sets of distorted images. Table 1 presents the values of Pearson correlation coefficient for the considered 6 metrics and the subsets used in LIVE database. Similarly, Table 2 contains the corresponding values of Spearman rank order correlation.

In terms of accuracy we can see that distortion measures derived from absolute deviation and relative deviation outperforms the ones based on the absolute value of the difference. Considering measures based on one parameter $Q_3$ is the most efficient while $Q_1$ is the worst. Surprisingly enough these results holds for every type of distortion under study. In other words the scale parameter shows the best correspondence to the perceived image quality. If we compare the most performing single parameter measure $Q_3$ to WNSIM except for one set of JPEG distorted images and noisy images the later show a higher degree of correlation with Human Visual System scores. This is no more the case when both parameters are used in the distortion measure. The results suggest that $Q_5$ is globally more performing than WNSIM. Note this is not the case for two sets of distorted image sets (JPEG2, JP2(2)), but differences may not be statistically significant in this cases.

**Table1.** Prediction accuracy for the quality measures using LIVE database.

| Dataset | JP2(1) | JP2(2) | JPEG1 | JPEG2 | Noise | Blur | Error |
|---|---|---|---|---|---|---|---|
| Q1 | 0.7427 | 0.6538 | 0.7843 | 0.6820 | 0.9357 | 0.7292 | 0.8017 |
| Q2 | 0.5353 | 0.6360 | 0.7673 | 0.4514 | 0.7695 | 0.5661 | 0.8049 |
| Q3 | 0.8176 | 0.7929 | 0.8984 | 0.8099 | 0.9381 | 0.8025 | 0.8471 |
| Q4 | 0.6566 | 0.7768 | 0.8363 | 0.5909 | 0.8946 | 0.7372 | 0.8887 |
| Q5 | 0.9415 | 0.9458 | 0.9682 | 0.9438 | 0.9412 | 0.9400 | 0.9558 |
| WNISM | 0.9353 | 0.9490 | 0.8452 | 0.9695 | 0.8902 | 0.8847 | 0.9221 |

Let us evaluate the performances of the proposed measures, but this time in term of prediction monotonicity. The three best metrics producing the greatest correlations for each subset are marked in bold in table 2.

**Table2.** Prediction monotonicity for the quality measures using LIVE database.

| Dataset | JP2(1) | JP2(2) | JPEG1 | JPEG2 | Noise | Blur | Error |
|---|---|---|---|---|---|---|---|
| Q1 | 0.7423 | 0.6445 | 0.7272 | 0.6783 | **0.9429** | 0.6933 | 0.8047 |
| Q2 | 0.4882 | 0.6297 | 0.7979 | 0.4214 | 0.7859 | 0.5769 | 0.7986 |
| Q3 | **0.8219** | 0.7923 | **0.8443** | **0.8137** | **0.9455** | **0.7925** | 0.8541 |
| Q4 | 0.6093 | 0.7567 | 0.8584 | 0.5123 | 0.8979 | 0.7586 | **0.8777** |
| Q5 | **0.8834** | **0.9268** | 0.8194 | 0.7610 | **0.9407** | **0.9334** | **0.9422** |
| WNISM | **0.9298** | **0.9470** | **0.8332** | **0.8908** | 0.8699 | **0.9147** | **0.9210** |

As illustrated, the measure $Q_5$ provides in most cases a monotonicity over 90%, whereas the measures $(Q_1, Q_2, Q_3, Q_4)$ fall below 80% except for some few cases. Although the WNISM seems to maintain a good monotonicity over all distortions, the measure $Q_5$ demonstrates its efficiency for noise, blur and transmission errors perturbations with a higher correlation than WNISM.

## 7. Conclusion

In this paper we investigate the design of a RR objective perceptual image quality metrics based on the natural image statistic approach. The statistics of the images derived in the tetrolet domain are modeled using a BKF distribution. This transform have demonstrate its efficiency to capture the local geometric structures of an image. It provides relevant statistical features since its statistical proprieties change considerably as the distortion type changes. The BKF parameters are involved to characterize the peakedness of the tetrolet coefficients distribution. The evaluation of the quality prediction performance reveals that the $L^2$-based measure outperforms favorably the measures based on only one parameter (scale or shape parameter). Moreover, this measure outperforms the prominent WNSIM one. The feature extraction step ensures a low computational complexity as a reduced amount of information has to be transmitted from the sender to the receiver. As the KLD is a most suitable distance to quantify the difference between two PDFs our future work will focus in deriving a closed expression of the KLD for the BKF model. Another way of improvement to investigate is the integration the HVS frequency and orientation sensitivity to the distortion measures.